\documentclass[journal]{IEEEtran}
\usepackage{amsmath,amssymb,amsfonts}
\usepackage{booktabs}
\usepackage{multirow}
\usepackage{array}
\usepackage{url}
\usepackage{xcolor}
\usepackage{graphicx}
\usepackage[hidelinks]{hyperref}

\newcommand{\method}{DRA-SAM}
\newcommand{\look}{Referential Disagreement LOOK}
\newcommand{\iou}{\operatorname{IoU}}
\newcommand{\mIoU}{\operatorname{mIoU}}
\newcommand{\Prec}[1]{\ensuremath{\text{\normalfont P@}#1}}
\newcommand{\figplaceholder}[1]{\fbox{\parbox[c][42mm][c]{0.92\linewidth}{\centering\small #1}}}
\newcommand{\includefigure}[2]{\IfFileExists{#1}{\includegraphics[width=\linewidth]{#1}}{\figplaceholder{#2}}}

\begin{document}

\title{DRA-SAM: Disagreement-Driven Referential Active SAM for Low-Altitude Drone Referring Segmentation}
\author{Jisheng~Dang, Huaiyuan~Peng, Wencan~Zhang, Bimei Wang, Jingwen~Zhao,
Wenjun Ma, Hong~Peng,
Bin~Hu,~\IEEEmembership{Fellow, IEEE}, Qi~Tian,~\IEEEmembership{Fellow, IEEE}, Tat-Seng~Chua%
\thanks{This work was supported in part by the National Natural Science Foundation of China under Grants 62227807 and U24B20186, in part by the Scientific and Technological Innovation 2030 Project of MOST, China, under Grants 2021ZD0200600 and 2021ZD0200408, and by the Supercomputing Center of Lanzhou University.}%
\thanks{Jisheng Dang, Huaiyuan Peng,  Bimei Wang, Jingwen Zhao, Wenjun Ma, Hong Peng, and Bin Hu are with the School of Information Science and Engineering, Lanzhou University, Lanzhou 730000, China.}%
\thanks{Wencan Zhang and Tat-Seng Chua are with the School of Computing, National University of Singapore, Singapore 119077.}%
\thanks{Qi Tian is with Huawei, Shenzhen 518129, China. 
Corresponding author: Bimei Wang and Bin Hu (e-mail: bh@bit.edu.cn).}}
\maketitle

\begin{abstract}
Referring segmentation in low-altitude drone imagery requires distinguishing small, visually similar instances under complex linguistic and spatial constraints. Global inference may yield multiple plausible masks or leave the intended referent insufficiently represented, motivating selective local re-inference. We propose DRA-SAM, a disagreement-driven framework that allocates an additional inference pass according to competing referential hypotheses and an unresolved state. A lightweight EvidenceScorer estimates beliefs over existing candidates and the unresolved state. A training-free Referential Disagreement LOOK policy balances belief-weighted candidate competition and unresolved-region coverage to select a crop for local re-inference. The same benchmark-adapted SAM3-I backbone processes the selected region, and the resulting evidence updates existing hypotheses and introduces new candidates. Among the compared methods, DRA-SAM achieves state-of-the-art performance on six of seven test metrics on both RIS-LAD and UAVid-RIS, including mIoU and \Prec{0.5}--\Prec{0.9}, and on all seven RIS-LAD validation metrics. On RIS-LAD, DRA-SAM improves the matched global SAM3-I baseline from 49.55 to 52.34 mIoU and from 46.80 to 58.05 \Prec{0.5}. On UAVid-RIS, the independently adapted DRA-SAM achieves 73.26 mIoU and 86.86 \Prec{0.5}. Controlled crop-policy comparisons demonstrate the effectiveness of disagreement-guided crop selection, while target-size analysis reveals particularly pronounced gains for aerial targets occupying less than 1\% of an image. These results establish referential disagreement as an effective criterion for allocating local inference. Code is available at \url{https://github.com/phy709/DRA-SAM}.
\end{abstract}


\section{Introduction}
\IEEEPARstart{R}{eferring} image segmentation (RIS) identifies the pixels of an object specified by a natural-language expression~\cite{hu2016segmentation}. Beyond recognizing object categories, RIS must resolve attributes and spatial relations among individual instances. Low-altitude drone imagery is challenging because global resizing often makes targets small and visually similar objects crowd the scene. RIS-LAD~\cite{ye2026rislad} exemplifies these conditions. A query describing a vehicle relative to its neighbors may correspond visually to several candidate masks, although only one is the intended referent.

\begin{figure*}[t]
\centering
\includefigure{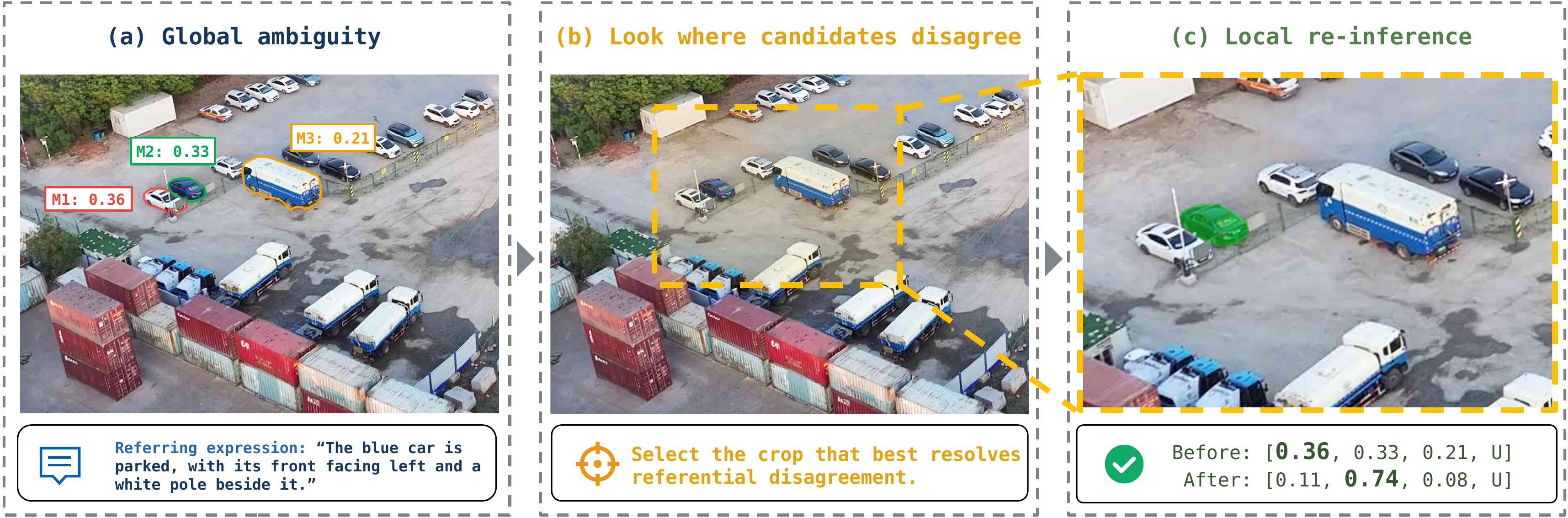}{Add Fig. 1 illustrating a low-altitude drone image with similar small objects, its referring expression, competing global masks, and the selected zoom-in crop.}
\caption{Motivation for active referring segmentation. (a) Global masks $M_1,M_2,M_3$ provide competing hypotheses. (b) DRA-SAM selects a crop covering referential disagreement or a potentially missing target. (c) Local SAM3-I inference updates the referent belief. $U$ denotes the unresolved hypothesis, not a numerical score in this schematic.}
\label{fig:intro}
\end{figure*}

Prior work advances RIS through cross-modal interaction and instance-aware reasoning~\cite{ye2019cmsa,ding2021vlt,liu2023isfp,cho2024crossvlt,liu2023m3att}, refined word--pixel alignment~\cite{zhang2022coupalign}, and scale-aware remote-sensing representations~\cite{yuan2024rrsis,liu2024rmsin}. Uncertainty-guided approaches such as CroBIM-U~\cite{sun2026crobimu} additionally regulate fusion and local refinement in spatially unreliable regions. Complementing these advances, we investigate \emph{where to allocate a new inference pass} to distinguish competing referents.

As illustrated in Fig.~\ref{fig:intro}, the global prediction gives rise to two complementary forms of referential ambiguity. \emph{Competition} occurs when several existing masks remain plausible and additional spatial detail can distinguish them. \emph{Discovery} concerns a referent that is insufficiently represented among the initial proposals and may become identifiable after local re-inference. These situations call for hypothesis-conditioned crop placement rather than reliance on a single global uncertainty score. By processing a selected image region at the backbone's input resolution, local re-inference increases the effective representation of small targets while preserving the context needed to interpret relational expressions. The added pass reprocesses pixels from the original image.

These two cases motivate a unified decision process. Candidate evidence identifies actionable disagreement, while the unresolved hypothesis reserves belief mass for potentially missing referents. Combining both signals allows one local prediction to refine an existing choice or introduce a missing candidate. Hypothesis-conditioned inference allocation is therefore central to referent resolution.
Therefore, we  propose \method{}, a disagreement-guided local re-inference framework based on SAM3-I~\cite{li2026sam3i}. A 382,920-parameter EvidenceScorer assigns beliefs to concrete mask candidates and an unresolved hypothesis. A parameter-free \look{} policy chooses a crop using belief-weighted mask disagreement (competition) and unresolved-region coverage (discovery). The same benchmark-adapted SAM3-I checkpoint predicts masks in this crop; these predictions update existing evidence and can introduce new candidates.

We evaluate DRA-SAM on RIS-LAD and UAVid-RIS with separate backbone adaptation. The benchmark comparisons show leading test-set mIoU and referring precision across both datasets, as well as the highest values on all seven RIS-LAD validation metrics. Beyond these cross-method comparisons, the matched-backbone experiment raises RIS-LAD mIoU from 49.55 to 52.34 and \Prec{0.5} from 46.80 to 58.05. Controlled crop-policy comparisons isolate the benefit of disagreement-driven selection. A target-size breakdown further shows \Prec{0.5} gains of 11.70--15.70 points for RIS-LAD objects covering less than 1\% of the image, directly connecting spatially selective inference with the small-object setting.

Our contributions are threefold:
\begin{itemize}
    \item We formulate active local re-inference over explicit mask hypotheses and an unresolved state, covering both candidate competition and potential target discovery.
    \item We develop a training-free disagreement-guided crop utility and a local-evidence update that re-ranks existing candidates and incorporates newly predicted masks.
    \item We validate on two UAV referring benchmarks using matched-backbone controls, crop-policy and utility ablations, and target-size analysis.
\end{itemize}

\section{Related Work}
\subsection{Cross-Modal and Instance-Aware Referring Segmentation}
RIS has evolved from recurrent language-conditioned segmentation~\cite{hu2016segmentation,li2018rrn} to long-range cross-modal attention~\cite{ye2019cmsa}, Transformer-based referring~\cite{ding2021vlt,yang2022lavt,kim2022restr}, and CLIP-assisted alignment~\cite{radford2021clip,wang2022cris}. CrossVLT~\cite{cho2024crossvlt} integrates stage-wise vision--language interactions, while multi-modal mutual attention~\cite{liu2023m3att} promotes iterative feature exchange. Instance-specific propagation~\cite{liu2023isfp}, multi-level language--mask alignment~\cite{zhang2022coupalign}, and multi-modal instance reasoning~\cite{hu2023dmmi} further improve referent discrimination. Compositional referring~\cite{xu2023meta} and instruction-guided segmentation~\cite{lai2024lisa,chen2024sam4mllm} broaden the task setting. Rather than redesigning these representations, DRA-SAM allocates an additional prediction to locations identified by disagreements among concrete mask hypotheses.

\subsection{Remote-Sensing and Drone Referring Segmentation}
Aerial visual grounding~\cite{sun2022vg} and RRSIS~\cite{yuan2024rrsis} established the importance of language-conditioned localization under small-object scale, crowded layouts, and diverse viewpoints. Subsequent work explores dual-modality alignment~\cite{pan2024dual}, rotated multi-scale interaction~\cite{liu2024rmsin}, fine-grained correspondence~\cite{lei2025fianet}, adaptive feature selection~\cite{yue2024asda}, and scale-aware prompting~\cite{lv2026resap}. RIS-LAD~\cite{ye2026rislad} focuses on low-altitude referring scenes, and UAVid-RIS~\cite{li2025aeroreformer} offers a complementary aerial benchmark. Our approach complements these representation-oriented advances by choosing the location of an additional inference pass after the global candidate set is available. This distinction is relevant to low-altitude imagery, where changing the processing scale of a particular region can be as useful as refining the global feature representation.

\subsection{Uncertainty, Prediction Correction, and Adaptive Computation}
Uncertainty modeling~\cite{kendall2017uncertainties} can identify unreliable predictions. UniA~\cite{yang2025unia} uses uncertainty inference and affinity diversification in weakly supervised segmentation, whereas CroBIM-U~\cite{sun2026crobimu} guides cross-modal fusion and local refinement with spatial referring uncertainty. DRA-SAM operates at a different level: it measures disagreement between \emph{complete candidate masks}, chooses an explicit crop, and integrates the resulting local proposals. Iterative prediction correction~\cite{shi2021qrn} and selective visual computation~\cite{mnih2014ram,jaderberg2015stn,rao2021dynamicvit} provide complementary motivations. Our LOOK criterion converts normalized hypothesis beliefs and spatial overlap into an analytical action utility, complementing learned attention and risk maps. Temperature scaling also controls the sharpness of hypothesis beliefs~\cite{guo2017calibration}. We select the belief temperature on internal validation to support candidate selection.

\subsection{Promptable and Foundation Segmentation}
Mask2Former~\cite{cheng2022mask2former}, SAM~\cite{kirillov2023sam}, and SAM 2~\cite{ravi2025sam2} have advanced general-purpose mask prediction, while Grounding DINO~\cite{liu2024groundingdino} and ReferFormer~\cite{wu2022referformer} connect language with object-level understanding. SAM 3~\cite{carion2025sam3} and SAM3-I~\cite{li2026sam3i} extend prompt- and instruction-driven segmentation. We adapt SAM3-I separately to each benchmark and then freeze it for both global and local inference. Thus, observed improvements from the proposed stage can be studied without changing the segmentation backbone.

\section{Proposed DRA-SAM Method}
\label{sec:method}
\subsection{Problem Setup and Framework Overview}
\begin{figure*}[t]
\centering
\includefigure{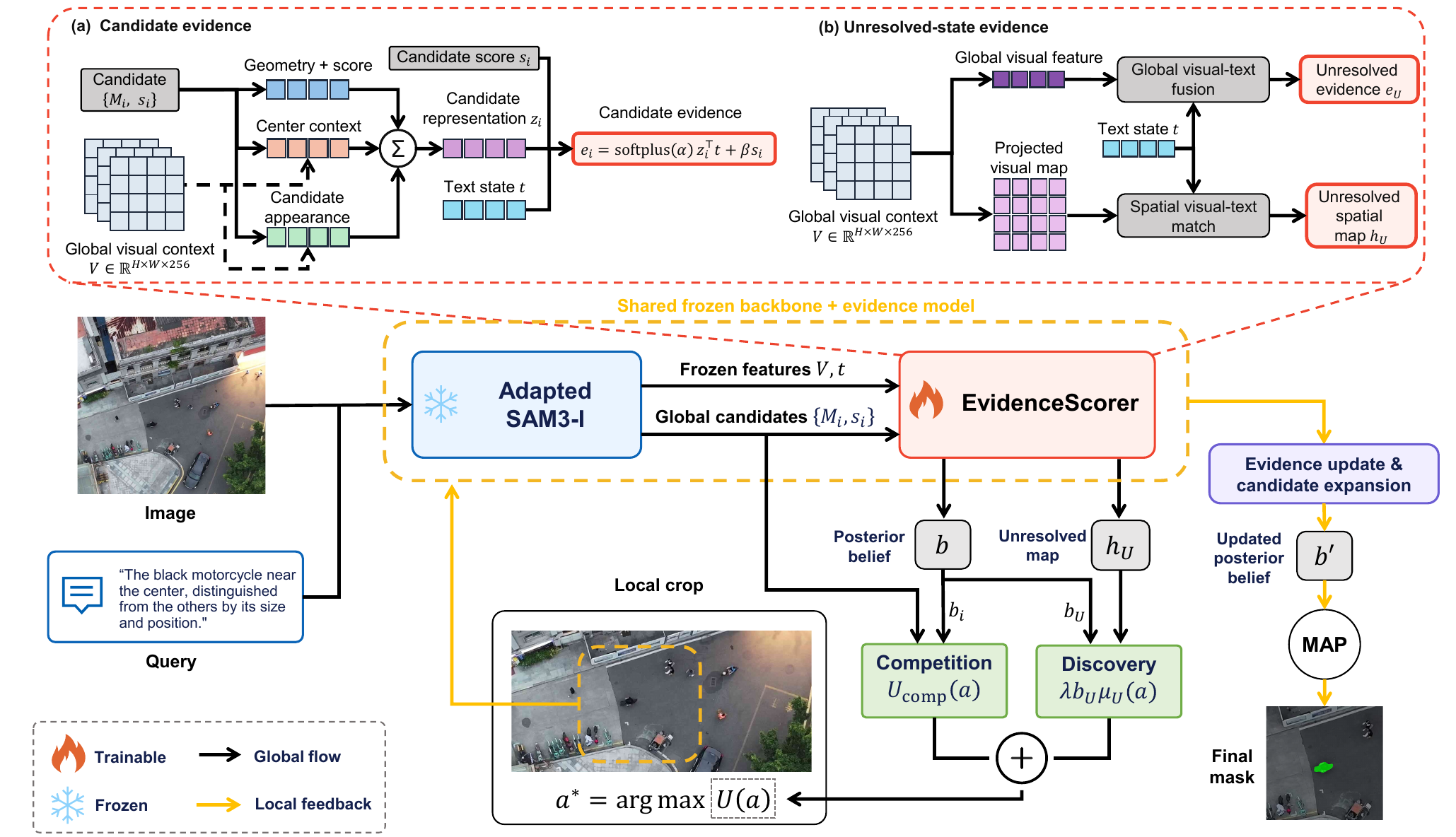}{Add Fig. 2 here: the overall DRA-SAM pipeline showing global inference with the adapted frozen SAM3-I backbone, EvidenceScorer-based initialization of the posterior, Referential Disagreement LOOK, local crop selection, additive evidence update, and final MAP prediction.}
\caption{Overview of DRA-SAM and its EvidenceScorer. Top: (a) mask-specific evidence $e_i$ from appearance, context, geometry, and language; (b) unresolved-state evidence $e_U$ and spatial prior $h_U$ from global visual--language features. Bottom: the frozen, benchmark-adapted SAM3-I generates global candidates; competition and discovery guide one local crop. Local evidence updates existing hypotheses and introduces new candidates, yielding beliefs $b'$ for final selection.}
\label{fig:method}
\end{figure*}

Given an image $x$ and expression $r$, the benchmark-adapted SAM3-I (Global) generates concrete masks
\begin{equation}
    \mathcal{M}_0=\{M_i\}_{i=0}^{K-1},
\end{equation}
with visual features $f_i$, normalized geometry $g_i=(c_i^x,c_i^y,w_i,h_i,\ldots)$, and raw scores. Index $0$ is a deterministic union-mask fallback. We append an unresolved hypothesis $U$ for a referent not adequately covered by the proposals and maintain the belief
\begin{equation}
    b=(b_0,\ldots,b_{K-1},b_U),\quad \sum_{i=0}^{K-1}b_i+b_U=1.
\end{equation}
A crop action $a=(a_x,a_y,s)$ specifies its normalized center and square scale. The action set $\mathcal{A}$ contains $24\times24$ centers at $s\in\{0.30,0.45,0.60,0.75\}$ (2,304 actions); $v_i(a)\in[0,1]$ measures a candidate's soft crop visibility. As shown in Fig.~\ref{fig:method}, EvidenceScorer initializes the beliefs, training-free LOOK selects one crop, and the same frozen, benchmark-adapted SAM3-I backbone produces local masks for evidence integration. The action thus allocates additional inference resolution within the original image. Candidate hypotheses represent specific predicted masks rather than object classes, allowing two vehicles of the same category to receive separate beliefs. The unresolved hypothesis provides a distinct route for exploration when the candidate set inadequately represents the described object. Both routes meet in a single crop decision and a common evidence-update procedure.

\subsection{Candidate Evidence and Initial Belief}
EvidenceScorer (382,920 trainable parameters) assigns evidence both to observed mask candidates and to the possibility of a missing referent (Fig.~\ref{fig:method}). For each mask, we resize it to the frozen backbone's visual-feature resolution, normalize its spatial mass, and pool a 256-dimensional appearance descriptor. The union fallback uses the mean descriptor of full-image proposals, or global average pooling if no proposals exist. Masked mean pooling of non-padding language tokens, followed by LayerNorm, a $256\!\rightarrow\!256$ projection, and $\ell_2$ normalization, yields the text embedding $t$.

Candidate geometry is encoded by

\begin{equation}
    g_i=(c_i^x,c_i^y,w_i,h_i,z_i^{\rm src},s_i^{\rm crop},u_i^{\rm union})\in\mathbb{R}^{7},
\end{equation}
where position and size are image-normalized, $z_i^{\rm src}$ marks crop-level origin, $s_i^{\rm crop}$ is crop scale (1 globally), and $u_i^{\rm union}$ marks the union fallback. Besides mask-pooled appearance, the scorer samples visual context at the mask center and embeds $[g_i;s_i]$ using the raw SAM3-I score $s_i$ (detection threshold 0.5). The three branches yield an $\ell_2$-normalized 256-dimensional candidate embedding $z_i$, without dropout or batch normalization. Its evidence combines cosine text affinity and the raw score:
\begin{equation}
    e_i = \operatorname{softplus}(\alpha)\, z_i^{\top}t + \beta s_i,
    \label{eq:candidate_evidence}
\end{equation}
The raw $\alpha,\beta$ parameters are initialized to 4.0 and 0.5; only the cosine scale is constrained to be positive. We retain all SAM3-I postprocessed masks, preceded by the union fallback, without further non-maximum suppression (NMS), deduplication, or top-$K$ filtering (RIS-LAD test: 1--9 initial candidates; median 3).

For unresolved evidence $e_U$, a bias-free $1\times1$ projection and normalized global visual pooling are concatenated with $t$, followed by Linear$(512,1)$ and a learned bias. Retaining evidence logits $\ell_j=e_j$, we compute
\begin{equation}
    b=\operatorname{softmax}\!\left(\frac{\ell}{\tau}\right),\quad
    \ell=(\ell_0,\ldots,\ell_{K-1},\ell_U).
\end{equation}
For discovery, the projected global map is also pooled to $24\times24$, channel-normalized, and compared with $t$ at each cell:
\begin{equation}
    h_U(p)=\operatorname{softmax}_{p}\!\left(
    \left\langle \frac{W_vV(p)}{\|W_vV(p)\|_2},t\right\rangle\right).
    \label{eq:unknown_map}
\end{equation}
Here, $h_U$ is a normalized crop-selection prior, rather than a foreground or final-mask prediction. The two outputs of EvidenceScorer play different roles: the vector $b$ specifies the relative support for candidate hypotheses, while $h_U$ identifies image regions compatible with the language expression. The unresolved belief $b_U$ determines how much the spatial prior influences crop choice. This links discovery to the current hypothesis state without a separate branch.

The crop policy is language-conditioned even though its search is analytical. Visual--text affinity shapes both the candidate beliefs $b_i$ and the spatial prior $h_U$, while $b_U$ scales the contribution of regions not adequately explained by existing masks. Consequently, different referring expressions can induce different spatial preferences in the same scene. A query distinguishing nearby vehicles may emphasize disagreement among plausible masks, whereas a description with stronger unresolved belief can prioritize regions indicated by $h_U$. LOOK thus uses cross-modal evidence to guide the location and scale of one additional visual observation.

\subsection{Disagreement-Driven LOOK Selection}
Belief dispersion alone cannot determine an informative crop. Two candidates may have similar scores yet overlap heavily, in which case their disagreement is localized differently from that of well-separated candidates. Conversely, a strongly ranked mask does not rule out a referent absent from the proposals. LOOK therefore combines \emph{competition} among existing masks with \emph{discovery} of unresolved targets, using only predicted evidence and no test-time ground truth. Pairwise mask disagreement is $D_{ij}=1-\iou(M_i,M_j)$. For $a=(a_x,a_y,s)$ and candidate $i$, define half-extents $\tilde h_{x,i}=\min(1,s/2+0.15w_i)$ and $\tilde h_{y,i}=\min(1,s/2+0.15h_i)$, and magnification $r(a)=\sqrt{0.30/\max(s,0.30)}$. Let $\mathcal{V}$ denote the valid candidate indices. A smooth geometric visibility proxy is
\begin{align}
    v_i(a)={}&r(a)\,\mathbf{1}_{\{i\in\mathcal{V}\}}\nonumber\\
    &\times\sigma\!\left(28(\tilde h_{x,i}-|c_i^x-a_x|)\right)\nonumber\\
    &\times\sigma\!\left(28(\tilde h_{y,i}-|c_i^y-a_y|)\right).
    \label{eq:visibility}
\end{align}
For a candidate pair, $Q_{ij}(a)=1-(1-v_i(a))(1-v_j(a))$ indicates whether at least one candidate is visible. The \emph{competition utility} becomes
\begin{equation}
    U_{\rm comp}(a)=\frac{1}{2}\sum_{i\neq j}b_ib_jD_{ij}Q_{ij}(a).
    \label{eq:competition}
\end{equation}
Allowing either candidate to be visible permits tight crops to test one hypothesis without enclosing both. This design differs from requiring both candidates to fit inside a crop, which can favor large regions and reduce the effective magnification of a small referent. The product $b_ib_j$ emphasizes pairs that remain plausible under the current belief, while $D_{ij}$ reflects their geometric difference. Consequently, competition rewards a crop that exposes a consequential candidate disagreement and can provide a discriminating new prediction.

For \emph{discovery}, we integrate $h_U$ over a soft crop window. Let $p=(x_p,y_p)$ index the $24\times24$ spatial grid and $h_a=s/2$:
\begin{equation}
\begin{aligned}
    \mu_U(a)={}&\sum_p h_U(p)\,
    \sigma\!\left(28(h_a-|x_p-a_x|)\right)\\
    &\cdot\sigma\!\left(28(h_a-|y_p-a_y|)\right),
\end{aligned}
    \label{eq:unknown_mass}
\end{equation}
where $\mu_U(a)$ is clipped to $[0,1]$. The complete crop utility is
\begin{equation}
    U(a)=U_{\rm comp}(a)+\lambda b_U\mu_U(a),\quad a^*=\arg\max_{a\in\mathcal{A}} U(a).
    \label{eq:utility}
\end{equation}
LOOK scores 2,304 candidate crop actions using predicted masks, beliefs, and geometry, without training a crop-policy network. Candidate competition focuses the crop on plausible, spatially distinct hypotheses, whereas unresolved belief directs it toward language-compatible regions. Importantly, position and scale are selected jointly through a resolution--coverage trade-off. Smaller crops magnify compact candidates when resized to the SAM3-I input resolution, and $r(a)$ favors this detailed examination in the competition term. Larger crops encompass more surrounding context and can capture greater discovery mass $\mu_U(a)$. The belief-weighted utility selects between discriminative detail and wider contextual coverage.

Action evaluation is separated from backbone inference. Competition uses candidate-pair disagreements and action-dependent visibilities, whereas discovery integrates the $24\times24$ spatial prior over each soft crop window. For $K$ concrete masks, a direct evaluation of all actions has pairwise complexity $O(|\mathcal{A}|K^2)$ and window-integration complexity $O(|\mathcal{A}|\cdot24^2)$. On RIS-LAD test, $K$ ranges from one to nine and $|\mathcal{A}|=2304$. The utility computations use predicted masks, candidate beliefs, and lightweight spatial features. Only the selected crop is forwarded through SAM3-I, keeping action scoring separate from the additional segmentation pass.

\emph{Geometric interpretation of scale.} Crops are resized to the same SAM3-I input resolution. When a referent lies fully within a crop of relative width $s$, its apparent linear extent increases approximately by $1/s$ compared with full-image resizing. The four scales $\{0.30,0.45,0.60,0.75\}$ thus provide nominal magnification factors of 3.33, 2.22, 1.67, and 1.33. Tighter views emphasize appearance details of small candidates, while wider views retain neighbors and their spatial relationships. The competition term favors magnified inspection through $r(a)$, and the discovery term favors coverage of language-compatible regions through $\mu_U(a)$. LOOK jointly selects location and scale from these complementary cues using a single local inference pass.

\subsection{Local Re-Inference and Evidence Integration}
The selected crop supplies evidence both for existing hypotheses and for new candidates. Given image dimensions $W,H$, its size is $\max(2,\operatorname{round}(Ws))\times\max(2,\operatorname{round}(Hs))$; the centered top-left coordinate is clamped to the image bounds. We insert each local binary mask into a full-image zero canvas and recompute its geometry, so all hypotheses share image coordinates.

Let $z_k^a$ denote a local candidate embedding, $c_i=(c_i^x,c_i^y)$ the normalized center, and $A_i$ the area of an existing hypothesis. Matching is soft and many-to-many, without an IoU threshold or a one-to-one assignment constraint. The pair score is
\begin{equation}
\begin{aligned}
    \pi_{ik}={}&\operatorname{softplus}(\gamma_1)z_i^\top z_k^a\\
    &-\operatorname{softplus}(\gamma_2)\lVert c_i-c_k^a\rVert_2\\
    &-\operatorname{softplus}(\gamma_3)\,\eta_{ik},
\end{aligned}
    \label{eq:pair_score}
\end{equation}
where $\eta_{ik}=\min(|\log(A_k^a/A_i)|,6)$ captures relative area. The learned raw $\gamma_1,\gamma_2,\gamma_3$ values start at 4.0, 2.0, and 0.5 and are constrained positive through softplus. For $n_a$ local candidates, matched evidence is
\begin{equation}
    m_i=\operatorname{logsumexp}_{k=1}^{n_a}(\pi_{ik})-\log n_a.
    \label{eq:matched_evidence}
\end{equation}
When no local candidate exists, we set $m_i=0$ before the miss rule. Define the miss indicator $\chi_i(a)=\mathbf{1}_{\{m_i<0\;\lor\;\text{no valid pair for }i\}}$. The existing-candidate update is
\begin{equation}
    \delta_i=v_i(a)\left[m_i-\chi_i(a)\operatorname{softplus}(\rho_{\rm miss})\right].
    \label{eq:existing_update}
\end{equation}
The miss parameter $\rho_{\rm miss}$ starts at 2.0, penalizing weak or absent evidence in proportion to crop visibility. We update retained raw logits as $\ell_i'=\ell_i+\delta_i$.

Each valid local mask also becomes a new hypothesis, without a proposal gate, duplicate merging, or extra NMS. Its logit combines existing-hypothesis support with unresolved mass:
\begin{equation}
\begin{aligned}
    \ell_k^{\rm new}={}&\log\!\Bigl[
    \sum_{i=0}^{K-1} \exp(\log b_i+\pi_{ik})\\
    &+\exp(\log b_U+\log\mu_U(a))\Bigr]+e_k^a,
\end{aligned}
\label{eq:new_candidate}
\end{equation}
Here, $e_k^a$ is local EvidenceScorer output and $\mu_U(a)$ is lower-bounded by $10^{-6}$ inside the logarithm. The residual unresolved logit becomes
\begin{equation}
    \ell_U'=\ell_U+
    \log\!\left(\max(1-\mu_U(a),10^{-6})\right).
    \label{eq:unknown_residual}
\end{equation}
We normalize concatenated existing, new, and unresolved logits using $b'=\operatorname{softmax}(\ell'/\tau)$. On RIS-LAD test, one crop adds 0--8 hypotheses (median 2), yielding 1--17 concrete candidates in total (median 4); no hard candidate cap is used. The integration supports two outcomes of the same local pass. Matching evidence adjusts the relative support for global masks whose appearance and geometry agree with the crop predictions; local masks are also admitted as concrete hypotheses, enabling target discovery beyond the original candidate pool. The unresolved logit decreases according to the explored spatial mass, so discovery evidence is reflected in the final belief distribution. The update integrates global and local evidence without assuming statistical independence.

\subsection{EvidenceScorer Supervision and Training}
We train EvidenceScorer on cached features from a fixed benchmark-adapted SAM3-I checkpoint (RIS-LAD Stage3; Section~\ref{sec:stage3_adaptation}). For ground-truth mask $G$, candidate quality is $q_i=\iou(M_i,G)$ and unresolved quality is $q_U=1-\max_i q_i$. At each supervision stage $\nu\in\{\mathrm{prior},\mathrm{post}\}$, soft-target beliefs are $y_j^{(\nu)}=\operatorname{softmax}(q_j^{(\nu)}/\tau_{\rm train})$, with $\tau_{\rm train}=0.10$. We jointly supervise initial and updated beliefs:
\begin{equation}
\begin{aligned}
    \mathcal{L}&=\frac{2}{3}\mathcal{L}_{\rm prior}+\frac{4}{3}\mathcal{L}_{\rm post},\\
    \mathcal{L}_{\nu}&=-\sum_j y_j^{(\nu)}\log b_j^{(\nu)}.
\end{aligned}
    \label{eq:training_loss}
\end{equation}
AdamW training uses 30 epochs, learning rate $2\times10^{-4}$, weight decay 0.05, batch size 4, one warmup epoch followed by cosine decay, gradient norm clipping at 1.0, bfloat16 (BF16) mixed precision, and seed 123. We use final-epoch weights for evaluation. On RIS-LAD, the feature cache contains 8,755 training and 957 internal-validation records; the separate official 1,396-sample validation split is reserved for reporting.

Local training observations are generated offline using the same frozen checkpoint, with up to 22 deduplicated crops per annotation: eight ground-truth-centered crops at scales $\{0.30,0.45,0.60,0.75\}$ and two jitter settings, eight shifted crops at the two smallest scales, and six random crops. Ground truth is used only for offline crop sampling and soft-target supervision. At test time, LOOK evaluates the full 2,304-action grid without ground-truth input. Because the cache omits full candidate masks, training-time LOOK uses normalized box-IoU disagreement; inference uses predicted-mask IoU. The EvidenceScorer and matching parameters are learned, whereas crop selection remains analytical.

\subsection{Final Referent Prediction}
After one local update, we select the highest-belief concrete mask. If the unresolved belief strictly exceeds every concrete belief, the deterministic union fallback (index 0) is returned. Confidence and risk gates are disabled:
\begin{equation}
    \hat{i}=\begin{cases}
    \arg\max_{0\le i<K'}b_i',&\max_{0\le i<K'}b_i'\ge b_U',\\
    0,&b_U'>\max_{0\le i<K'}b_i',
    \end{cases}
\end{equation}
where $K'$ includes the original and newly appended concrete candidates, and $b'$ denotes the post-update beliefs.

\section{Experiments}
\label{sec:experiments}
\begin{table*}[!t]
\centering
\caption{Comparison on RIS-LAD (\%). Boldface indicates the highest reported score per metric. Published methods follow their original protocols.}
\label{tab:published}
\scriptsize
\setlength{\tabcolsep}{2.1pt}
\resizebox{0.95\linewidth}{!}{%
\begin{tabular}{lrrrrrrr|rrrrrrr}
\toprule
\multirow{2}{*}{Method} & \multicolumn{7}{c|}{Validation} & \multicolumn{7}{c}{Test}\\
\cmidrule(lr){2-8}\cmidrule(lr){9-15}
 & \Prec{0.5} & \Prec{0.6} & \Prec{0.7} & \Prec{0.8} & \Prec{0.9} & oIoU & mIoU & \Prec{0.5} & \Prec{0.6} & \Prec{0.7} & \Prec{0.8} & \Prec{0.9} & oIoU & mIoU\\
\midrule
LAVT (2022)~\cite{yang2022lavt} & 33.95 & 31.16 & 26.29 & 19.41 & 9.89 & 44.03 & 32.25 & 31.67 & 27.04 & 23.02 & 17.55 & 8.58 & 41.97 & 30.14\\
LGCE (2024)~\cite{yuan2024rrsis} & 28.22 & 24.07 & 19.99 & 15.11 & 7.88 & 41.72 & 27.68 & 26.38 & 21.82 & 17.55 & 13.32 & 7.06 & 40.75 & 26.17\\
ASDA (2024)~\cite{yue2024asda} & 35.74 & 29.44 & 23.14 & 13.54 & 4.01 & 38.70 & 35.46 & 33.19 & 26.85 & 20.16 & 12.78 & 3.94 & 37.53 & 33.33\\
VATEX (2025)~\cite{nguyen2025vatex} & 19.27 & 14.76 & 9.74 & 6.02 & 2.08 & 24.83 & 20.32 & 16.07 & 11.87 & 8.47 & 4.67 & 1.63 & 24.27 & 18.53\\
FIANet (2025)~\cite{lei2025fianet} & 42.91 & 37.61 & 32.59 & 26.43 & 14.83 & 45.24 & 39.61 & 40.21 & 35.22 & 29.93 & 23.71 & 13.21 & 43.39 & 37.44\\
RMSIN (2024)~\cite{liu2024rmsin} & 45.85 & 40.97 & 35.24 & 28.01 & 16.33 & 50.17 & 42.08 & 43.36 & 38.11 & 32.36 & 25.08 & 13.75 & 48.82 & 39.60\\
RSRefSeg (2025)~\cite{chen2025rsrefseg} & 46.35 & 43.48 & 39.76 & 35.17 & 26.58 & 50.04 & 43.42 & 44.73 & 40.21 & 36.30 & 30.91 & 21.46 & 47.71 & 41.16\\
CADFormer (2025)~\cite{liu2025cadformer} & 45.34 & 38.90 & 33.60 & 26.86 & 15.40 & 47.37 & 41.36 & 42.20 & 36.99 & 31.92 & 23.60 & 12.63 & 46.47 & 39.32\\
SAARN (2026)~\cite{ye2026rislad} & 47.06 & 43.12 & 38.25 & 31.81 & 19.27 & 51.54 & 44.30 & 45.02 & 39.34 & 33.37 & 26.24 & 15.31 & 49.60 & 41.67\\
ReSaP-8B (2026)~\cite{lv2026resap} & 52.01 & 49.13 & 46.05 & 38.25 & 26.14 & 56.13 & 47.24 & 51.28 & 47.88 & 43.06 & 36.62 & 19.33 & 53.44 & 45.62\\
DeCo-L (2026)~\cite{lu2026deco} & -- & -- & -- & -- & -- & -- & -- & 56.10 & 51.60 & 45.90 & 35.50 & 18.90 & \textbf{54.80} & 50.20\\
\midrule
\textbf{DRA-SAM (Ours)} & \textbf{59.03} & \textbf{57.66} & \textbf{55.52} & \textbf{50.79} & \textbf{39.40} & \textbf{56.66} & \textbf{54.14} & \textbf{58.05} & \textbf{55.95} & \textbf{53.35} & \textbf{47.48} & \textbf{34.02} & 53.45 & \textbf{52.34}\\
\bottomrule
\end{tabular}}
\end{table*}

\subsection{Datasets, Metrics, and Implementation Details}
RIS-LAD~\cite{ye2026rislad} includes 9,712 training annotations (1,486 images), 1,396 validation and 2,763 test samples. A seed-123 image-disjoint split gives 8,755/957 scorer training/internal-validation records; hyperparameters are fixed before official evaluation. UAVid-RIS~\cite{li2025aeroreformer} includes 7,035 samples (3,215/1,163/2,657 train/validation/test). We use RIS-LAD for detailed ablations and UAVid-RIS for independent full-framework evaluation.

We report intersection over union (IoU) using its mean (mIoU) and overall (oIoU) variants, along with precision \Prec{\theta}, the percentage of samples whose prediction IoU is at least $\theta$. For predicted masks $P_n$ and references $G_n$ over $N$ samples, their aggregation differs:
\begin{equation*}
\begin{aligned}
\mIoU &=\frac{1}{N}\sum_{n=1}^{N}\frac{|P_n\cap G_n|}{|P_n\cup G_n|},\qquad
\mathrm{oIoU} =\frac{\sum_n|P_n\cap G_n|}{\sum_n|P_n\cup G_n|}.
\end{aligned}
\end{equation*}
Thus, mIoU gives each referred instance equal weight, whereas oIoU weights its IoU by its union area. Thresholded precision highlights the fraction of correctly localized and segmented instances at increasing overlap requirements. Together, these metrics characterize instance-level referring reliability and aggregate pixel overlap. A ground-truth-selected candidate oracle, when discussed, serves only as an upper bound and is never used at inference.

\subsubsection{Benchmark-Specific Backbone Adaptation and Fixed-Inference Protocol}
\label{sec:stage3_adaptation}
We first adapt SAM3-I on each benchmark's training split, then freeze its checkpoint for global inference, scorer training, and the additional local pass. For RIS-LAD, the final \emph{Stage3} referring-adaptation stage (not an epoch number) activates concept, simple-query, and complex-query branches with Kullback--Leibler (KL), enhanced-segmentation, InfoNCE, and margin objectives. The adapter and query/concept-projection modules are optimized, while the visual and language backbones remain frozen; this is neither full-model fine-tuning nor LoRA. Starting from the RIS-LAD Stage1/2 continuation checkpoint, we train for four epochs with AdamW, weight decay 0.1, batch size 1, BF16, no gradient accumulation, and gradient clipping at 0.1. Learning rates are $2.56\times10^{-3}$ (Transformer), $8\times10^{-4}$ (visual-backbone group), and $1.6\times10^{-4}$ (language-backbone group); frozen backbone parameters receive no updates. We use the final Stage3 weights for every RIS-LAD variant.

The \emph{SAM3-I (Global)} baseline uses this fixed checkpoint without cropping, obtaining 49.5545 test mIoU (49.55 when rounded). EvidenceScorer-only re-ranking and DRA-SAM differ solely in the use of one additional local pass and subsequent evidence update. The RIS-LAD protocol uses the epoch-30 EvidenceScorer checkpoint, 2,304 crop actions at scales $\{0.30,0.45,0.60,0.75\}$, discovery weight $\lambda=1.0$, belief temperature $\tau=1.2$, and a SAM3-I detection threshold of 0.5. Initial-risk and confidence-gain gates are disabled.

For UAVid-RIS, SAM3-I is independently adapted on its 3,215 training samples; EvidenceScorer is then trained with this \emph{frozen UAVid-RIS checkpoint}. Neither component is transferred directly from RIS-LAD. Both global and local passes share the resulting UAVid-RIS weights, and results are measured on the 2,657-sample official test split. Comparison values are taken from AeroReformer~\cite{li2025aeroreformer}.

\begin{table*}[!t]
\centering
\caption{UAVid-RIS test comparison (\%). Published baselines follow AeroReformer~\cite{li2025aeroreformer}; boldface indicates the highest reported value in each column.}
\label{tab:uavid}
\footnotesize
\setlength{\tabcolsep}{4pt}
\begin{tabular*}{0.95\textwidth}{@{\extracolsep{\fill}}lrrrrrrr@{}}
\toprule
Method & \Prec{0.5} & \Prec{0.6} & \Prec{0.7} & \Prec{0.8} & \Prec{0.9} & oIoU & mIoU \\
\midrule
RIS-DMMI (2023)~\cite{hu2023dmmi} & 79.75 & 70.19 & 54.08 & 36.13 & 9.26 & 76.76 & 67.10 \\
LAVT (2022)~\cite{yang2022lavt} & 84.45 & 75.75 & 59.60 & 38.71 & 10.91 & 78.76 & 69.32 \\
LGCE (2024)~\cite{yuan2024rrsis} & 83.97 & 74.60 & 59.69 & 39.37 & 11.33 & 79.06 & 69.52 \\
RMSIN (2024)~\cite{liu2024rmsin} & 85.71 & 77.91 & 64.06 & 46.52 & 17.76 & 81.10 & 72.05 \\
ASDA (2024)~\cite{yue2024asda} & 78.55 & 69.51 & 56.38 & 37.67 & 10.05 & 76.59 & 67.17 \\
MAFN (2025)~\cite{shi2025mafn} & 85.10 & 77.76 & 63.68 & 44.26 & 16.33 & 80.67 & 71.64 \\
AeroReformer (2025)~\cite{li2025aeroreformer} & 86.34 & 79.07 & 65.60 & 47.12 & 18.52 & \textbf{81.53} & 72.79 \\
\midrule
\textbf{DRA-SAM (Ours)} & \textbf{86.86} & \textbf{79.60} & \textbf{66.84} & \textbf{48.48} & \textbf{20.14} & 79.71 & \textbf{73.26} \\
\bottomrule
\end{tabular*}
\end{table*}

\subsection{Main Comparison on RIS-LAD}
Table~\ref{tab:published} compares DRA-SAM with reported methods on RIS-LAD. DRA-SAM ranks first among the compared methods on \emph{all seven validation metrics and six of seven test metrics}. On the test set, its 52.34 mIoU exceeds the next highest listed value, 50.20 from DeCo-L, by 2.14 points. Its \Prec{0.9} reaches 34.02, a 12.56-point gain over the strongest listed competing value of 21.46. The model also achieves the highest precision at every reported threshold, demonstrating gains across both permissive and stringent mask-overlap criteria. Test oIoU is 53.45, with DeCo-L reporting 54.80. Published baseline scores use the cited studies' evaluation protocols.

To establish the contribution of our inference design more directly, we additionally evaluate a \emph{matched} global SAM3-I checkpoint (Table~\ref{tab:components}). Relative to this backbone, DRA-SAM improves validation mIoU/oIoU/\Prec{0.5} from 51.09/51.40/49.79 to 54.14/56.66/59.03 and test scores from 49.55/49.08/46.80 to 52.34/53.45/58.05. This complementary comparison keeps backbone adaptation fixed and isolates the contribution of hypothesis scoring and selective re-inference.

\subsection{Main Comparison on UAVid-RIS}
Table~\ref{tab:uavid} evaluates DRA-SAM on a complementary UAV distribution with independently adapted SAM3-I and EvidenceScorer checkpoints. Among the compared methods, DRA-SAM ranks first in test mIoU and all five referring-precision thresholds (\Prec{0.5}--\Prec{0.9}), accounting for six of seven reported metrics. Relative to AeroReformer, its mIoU improves from 72.79 to 73.26 and \Prec{0.5} from 86.34 to 86.86. The precision advantage is maintained at stricter thresholds: \Prec{0.7} rises from 65.60 to 66.84, \Prec{0.8} from 47.12 to 48.48, and \Prec{0.9} from 18.52 to 20.14. DRA-SAM records 79.71 oIoU, while AeroReformer obtains the largest reported value of 81.53. These results show that hypothesis-guided inference allocation also benefits a second aerial referring benchmark under published evaluation protocols.

\emph{High-overlap precision across benchmarks.} At the strict \Prec{0.9} threshold, DRA-SAM reaches 34.02 versus the strongest listed alternative's 21.46 on RIS-LAD, and 20.14 versus 18.52 on UAVid-RIS. These results complement the improvements in \Prec{0.5}, indicating that the targeted local pass benefits both referent identification and high-overlap mask prediction across independently adapted aerial benchmarks.

\begin{table}[!htbp]
\centering
\caption{RIS-LAD test ablation of global candidate re-ranking and one local re-inference pass (\%).}
\label{tab:components}
\footnotesize
\setlength{\tabcolsep}{4pt}
\begin{tabular*}{0.95\linewidth}{@{\extracolsep{\fill}}lccc@{}}
\toprule
Protocol & \Prec{0.5} & oIoU & mIoU \\
\midrule
SAM3-I (Global) & 46.80 & 49.08 & 49.55 \\
SAM3-I + EvidenceScorer & 55.56 & 51.65 & 51.06 \\
\midrule
DRA-SAM (Ours) & \textbf{58.05} & \textbf{53.45} & \textbf{52.34} \\
\bottomrule
\end{tabular*}
\end{table}

\subsection{Contributions of Candidate Re-ranking and Local Re-Inference}
\label{sec:reinf_components}
Table~\ref{tab:components} progressively separates candidate re-ranking from local re-inference with the same adapted SAM3-I checkpoint. The global baseline directly uses the original predictions. EvidenceScorer-only selection instead compares the concrete masks and unresolved state without introducing new visual evidence, raising test mIoU/oIoU/\Prec{0.5} from 49.55/49.08/46.80 to 51.06/51.65/55.56. The full framework retains this scorer and adds one disagreement-selected local prediction, further improving all three scores to 52.34/53.45/58.05. The stepwise gains distinguish improvements from choosing among existing masks from those enabled by revisiting a spatial region and updating the candidate set. 

With the inference mechanism held fixed, we choose $\lambda=1.0$ from $\{0,0.25,0.5,0.75,1.0,1.25\}$ and $\tau=1.2$ from $\{0.8,1.0,1.2,1.5\}$ using the internal validation split. These yield 52.4340 mIoU and 57.1633 \Prec{0.5} for the selected $\lambda$, and 52.5158 mIoU, 54.0356 oIoU, and 57.1633 \Prec{0.5} for the selected $\tau$ in the corresponding sweeps. The settings are fixed before official testing.

\begin{figure}[!htbp]
\centering
\includegraphics[width=\columnwidth]{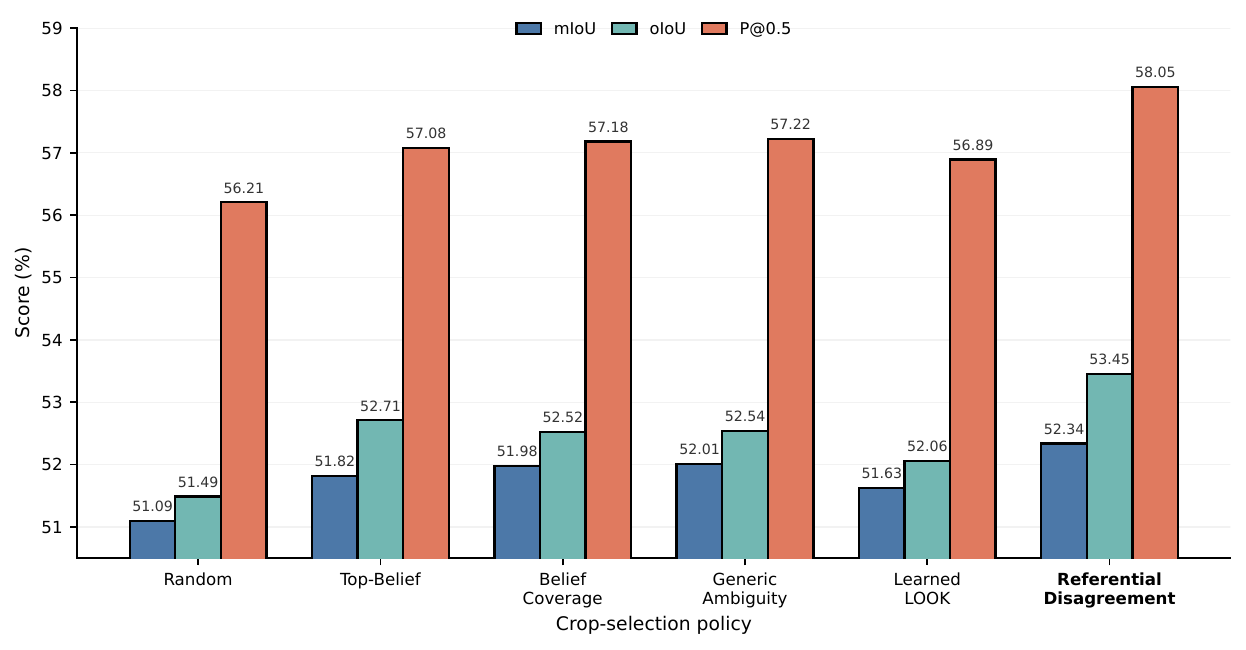}
\caption{RIS-LAD test comparison of six crop policies (\%). All policies share the same local-inference and evidence-update protocol; scores are annotated above the bars. The vertical axis is truncated for readability.}
\label{fig:look}
\end{figure}

\begin{figure*}[!t]
\centering
\includegraphics[width=\textwidth]{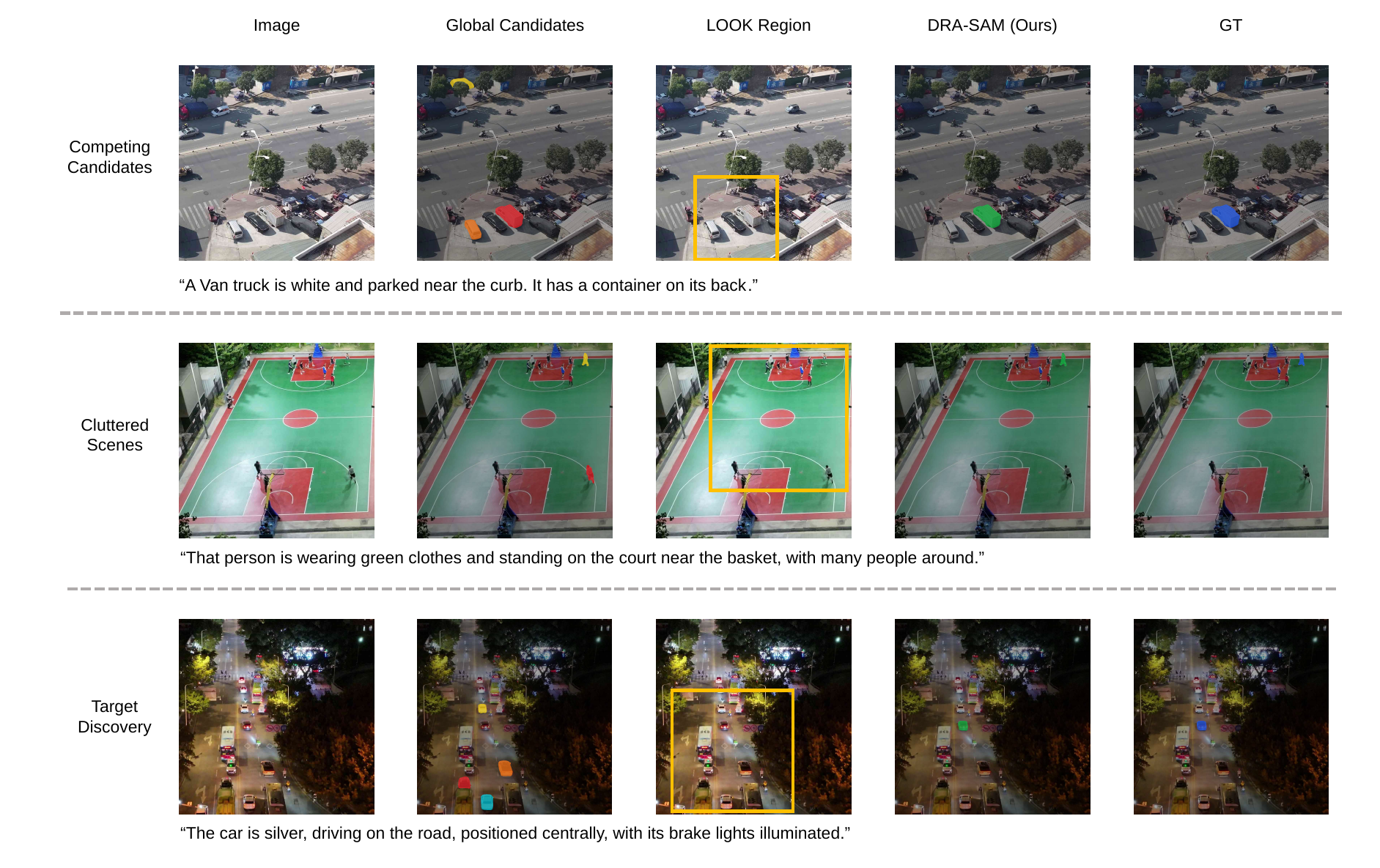}
\caption{Qualitative DRA-SAM results. Columns show the input, global candidates, selected LOOK crop (yellow), final prediction, and ground truth. Rows depict competing parked vehicles, a crowded basketball court, and a road-vehicle discovery scenario; referring queries appear beneath the images.}
\label{fig:qualitative}
\end{figure*}

\subsection{Controlled Comparison of LOOK Selection Policies}
To isolate spatial decision-making from the benefit of simply running the backbone twice, we compare six crop policies under identical action grids, local SAM3-I inference, and evidence updates. Random selects a crop without referential guidance. Top-Belief centers on the strongest candidate, Belief Coverage maximizes belief-weighted candidate coverage, and Generic Ambiguity uses a candidate-agnostic ambiguity score. A separately trained Learned LOOK supplies a learned-action alternative. Referential Disagreement instead places the action according to competition and discovery among explicit hypotheses.

Fig.~\ref{fig:look} shows that Referential Disagreement achieves the highest mIoU, oIoU, and \Prec{0.5} across the six policies. It exceeds Generic Ambiguity by 0.33, 0.91, and 0.83 points, respectively, and Random by 1.25, 1.96, and 1.84 points. The performance ordering supports the value of modeling \emph{which candidates disagree and where the unresolved evidence is located}, beyond generic uncertainty or belief coverage. Keeping the cost fixed at one local segmentation pass isolates crop selection from additional backbone computation.

\begin{table}[!htbp]
\centering
\caption{Crop-utility ablation on the RIS-LAD test set.}
\label{tab:utility}
\small
\begin{tabular*}{0.95\linewidth}{@{\extracolsep{\fill}}lccc@{}}
\toprule
Utility & \Prec{0.5} & oIoU & mIoU \\
\midrule
Competition only & 57.08 & 52.22 & 51.95 \\
Unknown only & 57.87 & \textbf{53.78} & 52.18 \\
Competition + unknown & \textbf{58.05} & 53.45 & \textbf{52.34} \\
\bottomrule
\end{tabular*}
\end{table}

\subsection{Qualitative Analysis of Active LOOK}
\label{sec:qualitative}
Fig.~\ref{fig:qualitative} traces the selected LOOK crop and final segmentation in three scenes, linking each crop to a distinct referring requirement. In the parked-vehicle example, similar candidates are close in appearance and the crop covers the relevant vehicle group, allowing the described van to be distinguished. The basketball scene contains crowded human instances and a relational cue involving the basket; the selected region retains the local configuration that makes this relation informative. In the roadway example, the additional crop examines a location from which a vehicle proposal can be recovered, illustrating the discovery branch. Together, the cases illustrate discrimination among existing proposals and expansion of the candidate set through one local pass. Crop selection relies on predicted evidence without test-time target masks.

\subsection{Contributions of Competition and Discovery Utilities}
Table~\ref{tab:utility} isolates the contributions of the two action-utility terms under the same inference protocol. Competition-only selection attends to areas associated with distinct, plausible concrete masks; discovery-only selection explores regions favored by the unresolved hypothesis. The combined policy reaches 52.34 mIoU and 58.05 \Prec{0.5}, the highest values among the three variants. Relative to competition alone, adding discovery improves mIoU by 0.39 and \Prec{0.5} by 0.97 points. Compared with discovery alone, the complete utility adds 0.16 mIoU and 0.18 \Prec{0.5} points. The discovery-only configuration attains the highest oIoU of 53.78. Both mechanisms contribute to referential accuracy through distinct spatial evidence.

\begin{table}[!htbp]
\centering
\caption{Oracle candidate-set analysis before and after one LOOK pass. Oracle mIoU and coverage are percentages; corresponding gains are percentage points.}
\label{tab:oracle}
\footnotesize
\setlength{\tabcolsep}{3pt}
\begin{tabular*}{0.95\linewidth}{@{\extracolsep{\fill}}lccc@{}}
\toprule
Candidate set & Mean $K$ & Coverage@0.5 & Oracle mIoU \\
\midrule
Global & 2.77 & 75.64 & 67.55 \\
Global + LOOK local masks & 4.57 & \textbf{78.97} & \textbf{70.71} \\
\midrule
Gain & +1.80 & +3.33 & +3.16 \\
\bottomrule
\end{tabular*}
\end{table}

\subsection{Oracle Analysis of Candidate Set Quality}
\label{sec:oracle}
We evaluate the quality of the mask hypothesis set by taking the best candidate IoU with ground truth for each sample. Table~\ref{tab:oracle} reports the resulting oracle mIoU and candidate coverage before and after one LOOK-selected local pass. The mean oracle IoU rises from 67.55\% to 70.71\%, and oracle coverage at IoU $\geq 0.5$ increases from 75.64\% to 78.97\%. After LOOK, 13.67\% of samples lacking a qualifying global candidate gain a qualifying local candidate. With 1.80 new masks per image on average, the local pass increases both the attainable segmentation overlap and the fraction of targets represented by a qualifying proposal. Together with Table~\ref{tab:components}, these gains quantify the discovery value of local inference.

On the 2,763-sample RIS-LAD test split, these oracle coverage rates correspond to approximately 2,090 instances with a qualifying global mask and 2,182 after incorporating local masks, a net gain of 92. At the final-mask level, EvidenceScorer re-ranking improves test \Prec{0.5} from 46.80 to 55.56, and LOOK-based local evidence further raises it to 58.05, as reported in Table~\ref{tab:components}. The combined evidence shows how active inference both expands the available hypotheses and improves the selected referent. Oracle labels are used only for evaluation; deployment selects from the learned evidence.

\begin{figure}[!htbp]
\centering
\includegraphics[width=\columnwidth]{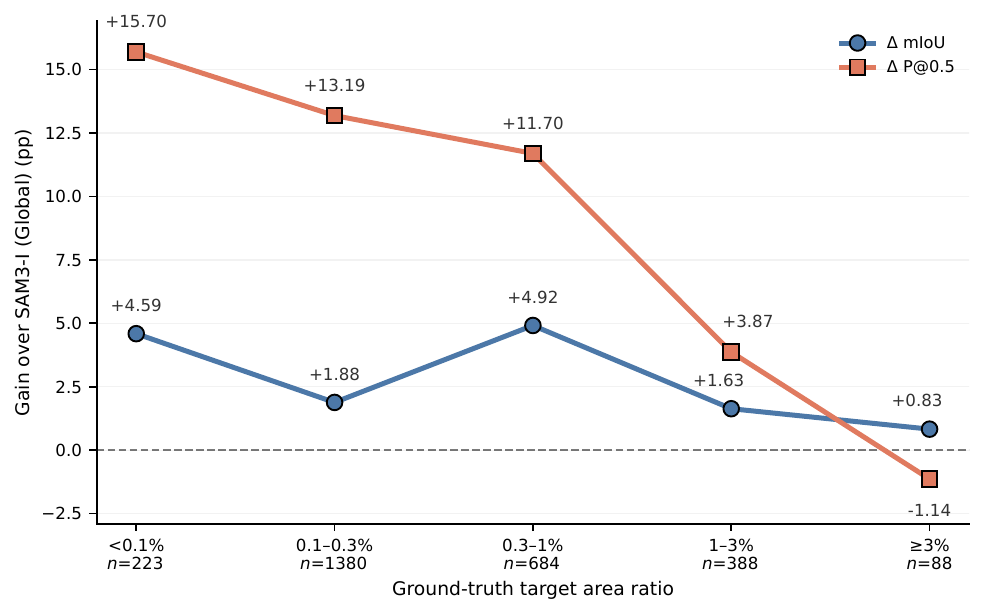}
\caption{RIS-LAD test gains over global SAM3-I by ground-truth area ratio $r_G=|G|/(HW)$ (percentage points). Sample counts appear beneath each bin. The largest \Prec{0.5} improvements occur below 1\% target area.}
\label{fig:size}
\end{figure}

\subsection{Target-Size Analysis of Local Re-inference}
Target size is a central characteristic of aerial RIS. Fig.~\ref{fig:size} partitions all 2,763 RIS-LAD test samples by the ground-truth area fraction $r_G=|G|/(HW)$. The smallest three groups contain 223, 1,380, and 684 samples, respectively; together, 2,287 of 2,763 examples (82.8\%) have targets smaller than 1\% of the image. Accurate small-instance referring is therefore central to this benchmark.
DRA-SAM improves mIoU in all five size groups, with gains of +4.59 points below 0.1\% and +4.92 points in the 0.3--1\% group. For the three groups below 1\%, \Prec{0.5} increases by 15.70, 13.19, and 11.70 points, respectively; the 1--3\% group gains 3.87 points. Gains are more concentrated on compact targets, consistent with reprocessing the chosen crop at greater effective target resolution. The final group (at least 3\%) records +0.83 mIoU, as also shown in Fig.~\ref{fig:size}.

\emph{Interpretation across IoU metrics.} The prominent small-target improvements also help explain the difference between the rankings under mIoU and oIoU. By definition, oIoU is a union-area-weighted average of per-sample IoU, so improvements on compact objects with small prediction--reference unions typically receive less weight than in mIoU. In contrast, mIoU and \Prec{\theta} count each referred instance equally. DRA-SAM's strong mIoU and threshold-precision results on both benchmarks, together with the size-stratified RIS-LAD gains, are therefore consistent with its objective of improved instance-level localization of small referents. The complementary metric trends reinforce the emphasis on recovering and accurately identifying small referred instances.

\subsection{Parameter Cost and Reproducibility}
EvidenceScorer contains 382,920 trainable parameters: 66,304 for text, 65,536 for visual projection, 66,304 each for candidate appearance and center context, 51,136 for geometry/score, 66,816 for candidate output, 513 for unresolved context, and seven learned scalars. SAM3-I remains frozen in this stage. At inference, one additional SAM3-I crop pass follows vectorized evaluation of 2,304 analytical actions.

Training comprises 65,670 optimization steps (2,189 batches per epoch for 30 epochs), with seed 123 and no gradient accumulation; training-action seeds additionally incorporate annotation identifiers. The pipeline separates cache extraction, scorer training, and end-to-end inference. Ground truth is restricted to offline crop sampling and IoU targets. Training-cache LOOK uses normalized box-IoU because masks are not stored, while test-time LOOK uses exact predicted-mask disagreement without ground-truth information.

\section{Conclusion}
We presented DRA-SAM, a hypothesis-conditioned framework that allocates one additional SAM3-I inference pass to a region selected from referential disagreement. EvidenceScorer models competition among concrete masks and the possibility of an unresolved target; training-free LOOK converts these beliefs into a crop action, and local predictions update or extend the hypothesis set. Across both datasets, DRA-SAM leads the reported comparisons in mIoU and all five referring-precision thresholds. Under a matched RIS-LAD backbone, it improves mIoU from 49.55 to 52.34 and \Prec{0.5} from 46.80 to 58.05. Controlled policy comparisons and scale-stratified analysis show the value of targeted inference allocation, particularly for small, crowded aerial referents. The framework complements global vision--language segmentation through targeted local inference.

\bibliographystyle{IEEEtran}
\bibliography{references}
\end{document}